\RequirePackage{fix-cm}
\documentclass[smallextended]{svjour3}       
\smartqed  
\usepackage{graphicx}
\begin{document}

\title{Neuroevolution in Deep Learning: The Role of Neutrality}


\author{Edgar Galv\'an}

\institute{Edgar Galv\'{a}n \at
 Naturally Inspired Computation Research Group, Department of Computer Science, National University of Ireland Maynooth, Ireland\\
              \email{edgar.galvan@mu.ie}
}

\date{Received: date / Accepted: date}

\maketitle

\begin{abstract}
  A variety of methods have been applied to the architectural configuration and learning or training of artificial deep neural networks (DNN). These methods play a crucial role in the success or failure of the DNN for most problems and applications. Evolutionary Algorithms (EAs) are gaining momentum as a computationally feasible method for the automated optimisation of DNNs. Neuroevolution is a term which describes these processes of automated configuration and training of DNNs using EAs. However, the automatic design and/or training of these modern neural networks through evolutionary algorithms is computanalli expensive. Kimura’s neutral theory of molecular evolution states that the majority of evolutionary changes at \sloppy{molecular} level are the result of random fixation of selectively neutral mutations. A mutation from one gene to another is neutral if it does not affect the phenotype. This work discusses how neutrality, given certain conditions, can help to speed up the training/design of deep neural networks . 
\end{abstract}

\section{Introduction}
\label{intro}
Deep learning algorithms, commonly referred as Deep Neural Networks~\cite{Goodfellow-et-al-2016,10.1162/neco.2006.18.7.1527,DBLP:journals/nature/LeCunBH15}, are inspired by deep hierarchical structures of human perception as well as production systems~\cite{galvan2020neuroevolution}. These algorithms have achieved expert human-level performance in multiple areas including computer vision problems~\cite{DBLP:conf/cvpr/SzegedyLJSRAEVR15},  games~\cite{Silver_2016}, to mention a few examples. The design of deep neural networks (DNNs) architectures (along with the optimisation of their hyperparameters) as well as their training plays a crucial part for their success or failure~\cite{LIU201711}.

Neural architecture search is a reality: a great variety of methods have been proposed over recent years including Monte Carlo-based simulations~\cite{negrinho2017deeparchitect}, random search~\cite{journals/jmlr/BergstraB12} and random search with weight prediction~\cite{DBLP:journals/corr/abs-1708-05344},  hill-climbing~\cite{elsken2017simple}, grid search~\cite{DBLP:conf/bmvc/ZagoruykoK16}, Bayesian optimisation~\cite{10.5555/3042817.3042832,DBLP:conf/nips/KandasamyNSPX18}, gradient-based~\cite{liu2018darts,xie2018snas}, and mutual information~\cite{Tapia_2020,DBLP:journals/corr/TishbyZ15,yu2019understanding}. However, two methods started gaining momentum thanks to their impressive results: reinforcement learning (RL) methods~\cite{10.5555/3312046}  and evolution-based methods~\cite{Back:1996:EAT:229867,EibenBook2003}, sometimes referred to as neuroevolution in the context of neural architecture search~\cite{galvan2020neuroevolution}, whereas the latter method started dominated the area due to better performance in e.g., terms of accuracy, as well as being reported to require less computational time to find competitive solutions~\cite{DBLP:conf/aaai/RealAHL19,8712430} compared to reinforcement learning methods.

\section{State-of-the-art in Neuroevolution in Deep Neural Networks}
\label{sec:state}

\begin{figure}[tbh!]
  \centering    
      \includegraphics[width=0.95\textwidth]{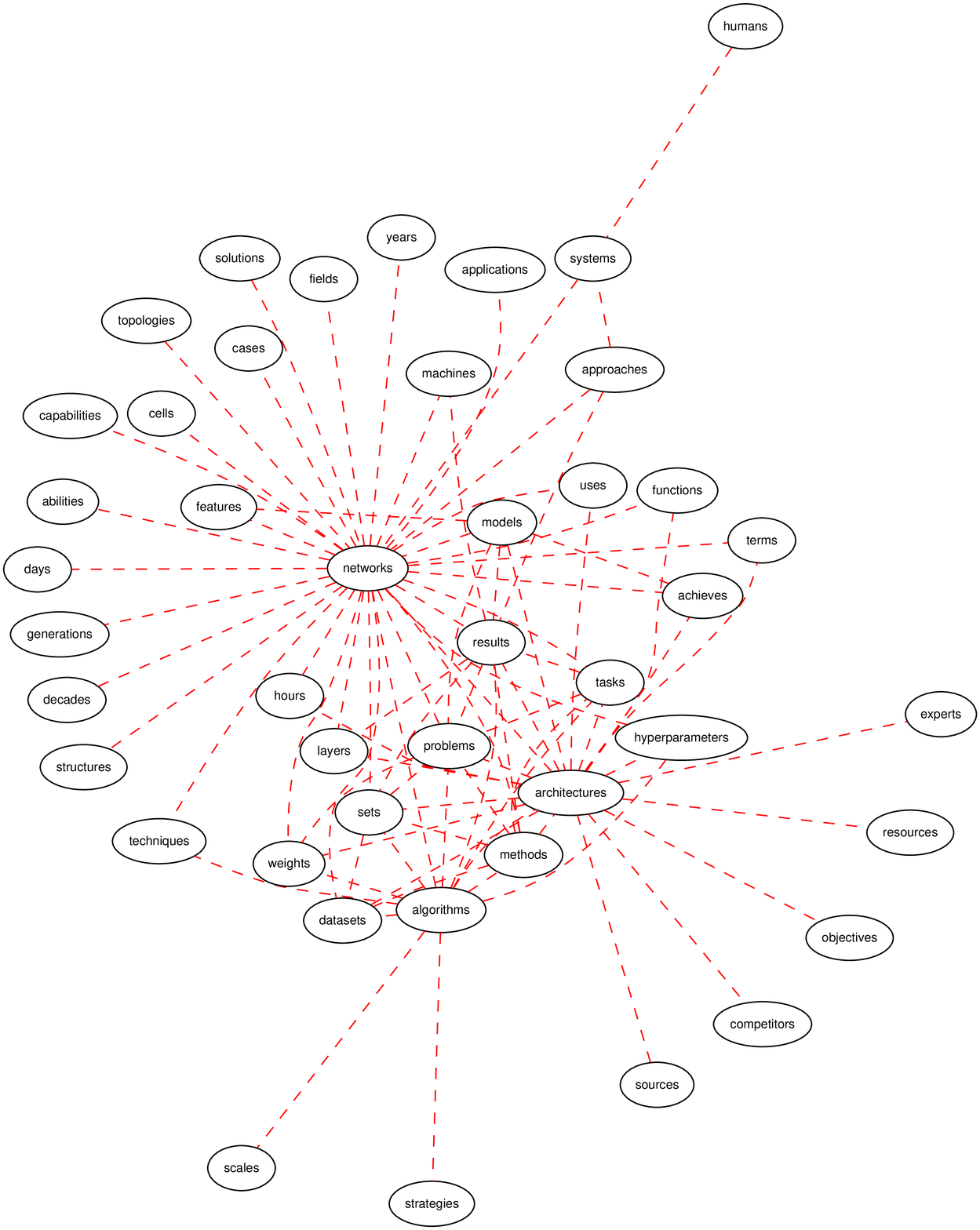}
\caption{Bird’s-eye view analysis of the research conducted in the area of neuroevolution in DNNS. The rest-length for repulsive forces between nodes was set to 13.}

\label{fig:birdsView}
\end{figure}

There has been an increased interest in the correct design (and to a lesser degree training) of deep neural networks by means of Evolutionary Algorithms, as extensively discussed in our recent work, summarising over 100 recent papers in the area or neuroevolution in deep neural networks~\cite{galvan2020neuroevolution}. Figure~\ref{fig:birdsView} shows a visual representation of the research trends followed in neuroevolution in deep neural networks. This is the result of using keywords used in titles and abstract of around 100 published in the last 5 years. We computed a similarity metric between these keywords and each paper. These similarities induce corresponding graph structures on the paper and key term `spaces’. Each paper/term corresponds to a node and edges arise naturally whenever there is a similarity between nodes. Details on how to generate this graph are given in~\cite{Poli:2008:APA:1362102.1384933}.

\subsection{Evolving Deep Neural Networks' Architectures with Evolutionary Algorithms}

The use of evolution-based methods in designing DNN is already a reality as discussed in~\cite{galvan2020neuroevolution}. Different Evolutionary Algorithms with different representations have been used, ranging from landmark evolutionary methods including  Genetic Algorithms~\cite{10.5555/531075}, Genetic Programming~\cite{Koza:1992:GPP:138936} and Evolution Strategies~\cite{10.1023/A:1015059928466,10.1007/978-3-642-81283-5_8} up to using hybrids combining, for example, the use of Genetic Algorithms and Grammatical Evolution~\cite{10.1007/BFb0055930}. In a short period of time, we have observed both ingenious representations and interesting approaches achieving extraordinary results against human-expert configured networks~\cite{DBLP:conf/aaai/RealAHL19}. We have also seen  state-of-the-art approaches in some cases employing hundreds of computers~\cite{10.5555/3305890.3305981} to using just a few GPUs~\cite{8712430}. Most  neuroevolution studies have focused their attention in designing deep Convolutional Neural Networks. Other networks have also been considered including Autoencoders, Restricted Boltzmann Machines, Recurrent Neural Networks and Long Short Term Memory, although there are just a few neuroevolution works considering the use of these types of networks. 

Our recent article~\cite{galvan2020neuroevolution} summaries, in a series of informative tables, the EA representation used, the representation of individuals, genetic operators used, and the EA parameters. They also outline the computational resources used in the corresponding study by attempting to outline the number of GPUs used. A calculation of the GPU days per run is approximated as in Sun et al.~\cite{8742788}. We indicate benchmark datasets used in the experimental analysis. Finally, the table indicates if the neural network architecture has been evolved automatically or by using a semi-automated approach whilst also indicating the target DNN architecture. Every selected paper does not report the same information. Some papers omit details about computational resources while others omit information about the number of runs. A very interesting output from this summary is that there are numerous differences between the approaches used by all of the papers listed. Crossover is omitted from several studies mostly due to encoding adopted by various researchers. Population size and selection strategies for the EAs change between studies. While our recent article~\cite{galvan2020neuroevolution} clearly demonstrates that  MNIST and CIFAR are the most popular benchmark datasets we can see many examples of studies using benchmark datasets from specific application domains.

\subsection{Training Deep Neural Networks Through Evolutionary Algorithms}

In the early years of neuroevolution, it was thought that evolution-based methods might exceed the capabilities of backpropagation~\cite{784219}. As Artificial Neural Networks, in general, and  as Deep Neural Networks (DNNs), in particular, increasingly adopted the use of stochastic gradient descent and backpropagation, the idea of using Evolutionary Algorithms (EAs) for training DNNs instead has been almost abandoned by the DNN research community. EAs are a ``genuinely different paradigm for specifying a search problem''~\cite{10.1145/2908812.2908916} and provide exciting opportunities for learning in DNNs. When comparing neuroevolutionary approaches to other approaches such as gradient descent, authors such as Khadka et al.~\cite{DBLP:journals/ec/KhadkaCT19} urge caution. A generation in neuroevolution is not readily comparable to a gradient descent epoch. Despite the fact that it has been argued that EAs can compete with gradient-based search in small problems as well as using NN with a non-differentiable activation function~\cite{MANDISCHER200287}, the encouraging results achieved in the 1990s~\cite{Goerick_evolutionstrategies:,10.5555/1623755.1623876,10.1109/64.393138} have inspired  some researchers to carry out research in training DNNs. This includes the work conducted by David and Greental~\cite{10.1145/2598394.2602287} and Fernando et al.~\cite{10.1145/2908812.2908890} both of which using deep autoeconders and Pawelczyk et al.~\cite{10.1145/3205651.3208763} and Such et al.~\cite{Such2017DeepNG} who use deep Convolutional Neural Networks. An informative summary of the works carried out on the training of DNNs using Evolutionary Algorithms can be seen in the our recent article~\cite{galvan2020neuroevolution}.

\section{Mutations and Neutral Theory}
\label{sec:mutations}

Kimura's neutral theory of molecular evolution~\cite{Kimura,kimura_1983} states that the majority of evolutionary changes at molecular level are the result of random fixation of \textit{selectively neutral mutations}. A mutation from one gene to another is neutral if it does not affect the phenotype. Thus, most mutations that take place in natural evolution are neither advantageous nor disadvantageous for the survival of individuals. It is then reasonable to extrapolate that, if this is how evolution has managed to produce the amazing complexity and adaptations seen in nature, then  neutrality should aid also EAs. However, whether neutrality helps or hinders the search in EAs is ill-posed and cannot be answered in general. One can only answer this question within the context of a specific class of problems, (neutral) representations and set of operators~\cite{DBLP:phd/ethos/GalvanLopez09,DBLP:conf/eurogp/LopezDP08,DBLP:conf/gecco/LopezP06,DBLP:conf/ppsn/LopezP06_2,DBLP:conf/micai/LopezP09,DBLP:journals/evs/LopezPKOB11,10.1007/978-3-540-73482-6_9,DBLP:journals/tec/PoliL12}.

We are not aware of any works in neuroevolution in DNNs on neutrality. In our recent in-depth review article on neuroevolution in deep neural networks~\cite{galvan2020neuroevolution}, we have seen that numerous studies used selection and mutation only to drive evolution in automatically finding a suitable deep neural network architecture or to train a neural network.  Interestingly, many researchers have reported highly encouraging results when using these two genetic operators, including the works conducted by Real et al.~\cite{DBLP:conf/aaai/RealAHL19,10.5555/3305890.3305981} using  GAs and hundreds of GPUs as well as the work carried out by Suganuma et al.~\cite{10.1145/3071178.3071229} employing Cartesian Genetic Programming and using only a few GPUs.

\begin{figure}[tbh!]
  \centering
      \includegraphics[width=0.95\textwidth]{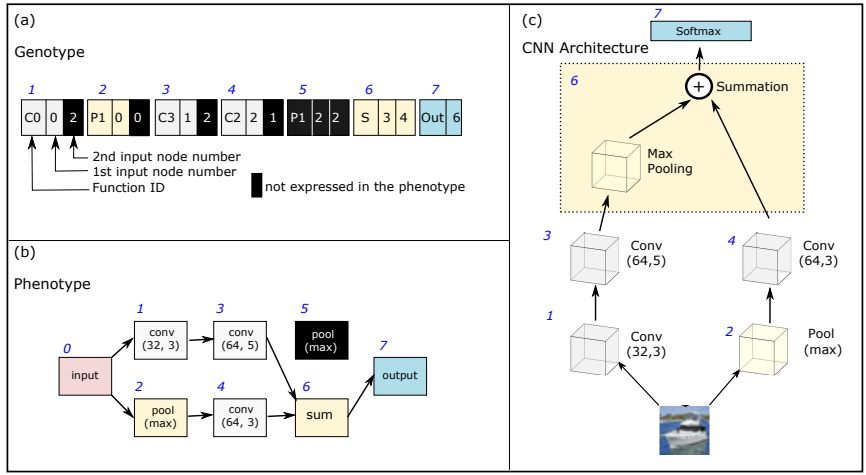}
\caption{\textbf{(a)} Genetic representation of a cartesian GP individual encoding a CNN architecture. \textbf{(b)} The phenotypic representation. \textbf{(c)}  CNN architecture defined by (a).  Gene No. 5, coloured with a black background  in the genotype (a) is not expressed in the phenotype. The summation node in (c),  with  light yellow background, performs max pooling to the LHS of the input (Node no. 3) to get the same input tensor sizes.  Redrawn from Suganuma et al.~\cite{10.1145/3071178.3071229}.}
\label{fig:cartesianCNN}
\end{figure}

If neutrality is beneficial, taking into consideration specific classes of problems, representations and genetic operators, this can also have an immediately positive impact in the  time needed to test the configuration of DNNs because the evaluation of potential EA candidate solutions will not be necessary. There are some interesting encodings adopted by researchers including Suganuma's work~\cite{10.1145/3071178.3071229} (see Fig.~\ref{fig:cartesianCNN}) that allow the measurement of the level of neutrality present in evolutionary search and can potentially indicate whether its presence is beneficial or not in certain problems and DNNs.

Fig.~\ref{fig:cartesianCNN} helps to illustrate how neutrality can be explicitly be promoted (or impeded) in evolutionary algorithms. The genotypic representation of a cartesian genetic programming~\cite{Miller2011} individual encoding a CNN architectures is shown in Fig~\ref{fig:cartesianCNN} (a). This is then decoded to a phenotypic representation Fig.~\ref{fig:cartesianCNN} (b), worth noting is how gene number 5 in the genotype is not expressed in the phenotype. Thus, any mutation taking place in gene 5  will not affect the phenotype which defines the CNN architecture depicted in Fig.~\ref{fig:cartesianCNN} (c).  

\subsection{Does neutrality help or hinder the search of an Evolutionary Algorithm?}

This question has been debated at considerable length in the literature without really reaching any form of consensus on its answer. The reasons for this situation include the lack of a single definition of neutrality, the multiple ways in which one can add neutrality to a representation, the focus on pure performance when evaluating the effects of neutrality without attention to the changes in the behaviour of the search operators and in the features of the fitness landscape, and, finally, the variability in the choice of problems, algorithms and representations for benchmarking purposes. Also, very often studies consider problems and representations that are quite complex and results represent the composition of multiple effects.

\section{First Research Steps in Neutrality in Neuroevolution in Deep Neural Networks}
\label{sec:first}

We believe that one of the first step to see whether neutrality helps or hinders evolution in the configuration (or training) of a deep neural network is to adopt a very simple representation such as binary representation, using mutation and selection as genetic operators to guide evolution. The type of problem is a more difficult endevour when trying to carry out this research. The reason is because much of the empirical scientific  works conducted in the area of neuroevolution in deep neural networks are incredible different, as summarised in our recent article~\cite{galvan2020neuroevolution}, where CNNs and computer vision datasets have been the attention of the research community and no general conclusions have been drawn in the area. However, these two can also represent good areas to be studied given the numerous results reported in a variety of studies, helping us to use them as basis for our research.

\bibliographystyle{abbrv}

\bibliography{neuroevolution}   

\end{document}